\documentclass[acmtog]{acmart}
\usepackage{cleveref}
\usepackage{algorithm}
\usepackage{subcaption}
\usepackage{stfloats}
\usepackage{multirow}
\usepackage{xspace}
\usepackage{color}

\AtBeginDocument{%
  }

\acmJournal{TOG}
    
\title{DiffCamera: Arbitrary Refocusing on Images}

\author{Yiyang Wang}
\orcid{0009-0000-1352-5883}
\affiliation{
 \institution{The University of Hong Kong}
 \country{Hong Kong}
 }
\email{yiyangwang@connect.hku.hk}

\author{Xi Chen}
\orcid{0009-0008-5008-4720}
\affiliation{
\institution{The University of Hong Kong}
\country{Hong Kong}
}
\email{xichen.csai@gmail.com}

\author{Xiaogang Xu}
\orcid{0000-0002-7928-7336}
\affiliation{
 \institution{The Chinese University of Hong Kong}
 \country{Hong Kong}
}
\email{xiaogangxu00@gmail.com}

\author{Yu Liu}
\orcid{0000-0001-8071-3745}
\affiliation{
\institution{Tongyi Lab}
\country{China}
}
\email{ly103369@alibaba-inc.com}

\author{Hengshuang Zhao}
\orcid{0000-0001-8277-2706}
\authornote{Corresponding author.}
\affiliation{
 \institution{The University of Hong Kong}
 \country{Hong Kong}
}
\email{hszhao@cs.hku.hk}

\newcommand{\methodname}{DiffCamera}
\newcommand{\method}{\text{\methodname}\xspace}

\begin{abstract}
The depth-of-field (DoF) effect, which introduces aesthetically pleasing blur, enhances photographic quality but is fixed and difficult to modify once the image has been created. This becomes problematic when the applied blur is undesirable~(\textit{e.g.}, the subject is out of focus).
To address this, we propose \method, a model that enables flexible refocusing of a created image conditioned on an arbitrary new focus point and a blur level.
Specifically, we design a diffusion transformer framework for refocusing learning. However, the training requires pairs of data with different focus planes and bokeh levels in the same scene, which are hard to acquire.
To overcome this limitation, we develop a simulation-based pipeline to generate large-scale image pairs with varying focus planes and bokeh levels.
With the simulated data, we find that training with only a vanilla diffusion objective often leads to incorrect DoF behaviors due to the complexity of the task.
This requires a stronger constraint during training.
Inspired by the photographic principle that photos of different focus planes can be linearly blended into a multi-focus image, we propose a stacking constraint during training to enforce precise DoF manipulation.
This constraint enhances model training by imposing physically grounded refocusing behavior that the focusing results should be faithfully aligned with the scene structure and the camera conditions so that they can be combined into the correct multi-focus image.
We also construct a benchmark to evaluate the effectiveness of our refocusing model.
Extensive experiments demonstrate that \method supports stable refocusing across a wide range of scenes, providing unprecedented control over DoF adjustments for photography and generative AI applications.
\end{abstract}

\ccsdesc[500]{Computing methodologies~Computer vision}

\keywords{Image editing, Depth-of-field}

\begin{teaserfigure}
  \includegraphics[width=\textwidth]{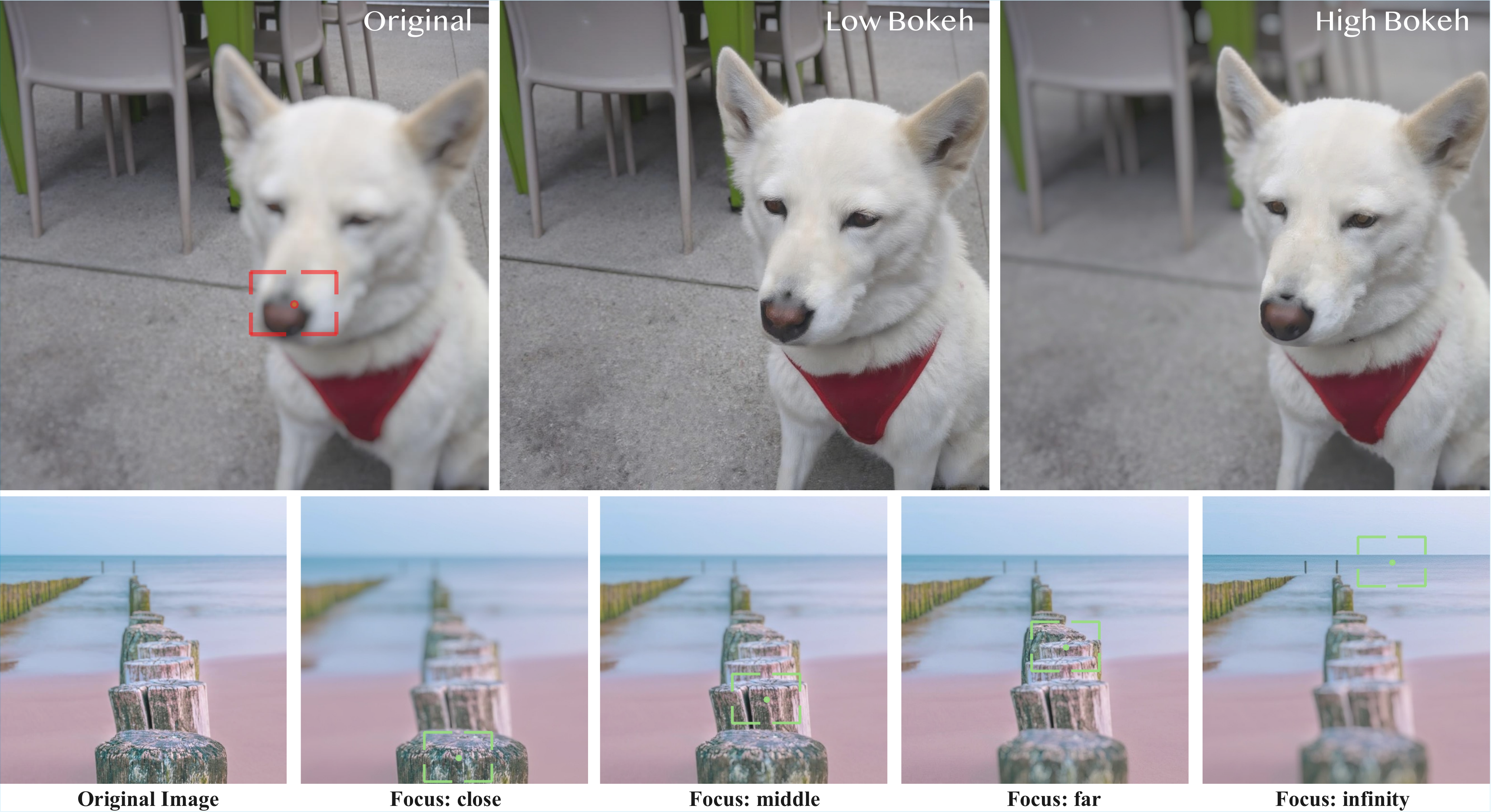}
  \caption{
  \textbf{Demonstrations of DiffCamera's refocus abilities.}
 \method enables refocusing an image on any specified focus point with a designated bokeh (blur) level, maintaining scene consistency regardless of the original depth-of-field (DoF) effects. The focus frame marks the target focus point.
  }
  \label{fig:teaser}
\end{teaserfigure}

\copyrightyear{2025}
\acmYear{2025}
\setcopyright{acmlicensed}\acmConference[SA Conference Papers '25]{SIGGRAPH Asia 2025 Conference Papers}{December 15--18, 2025}{Hong Kong, Hong Kong}
\acmBooktitle{SIGGRAPH Asia 2025 Conference Papers (SA Conference Papers '25), December 15--18, 2025, Hong Kong, Hong Kong}
\acmDOI{10.1145/3757377.3763827}
\acmISBN{979-8-4007-2137-3/2025/12}

\begin{document}
\maketitle

\section{Introduction}

Depth-of-field (DoF) offers a wide range of creative possibilities in photography. The intentional blurriness brought by DoF in photographs, known as bokeh effects, emphasizes specific subjects. By adjusting the DoF, photographers can selectively focus on subjects, control background blur, and enhance the visual impact of their images. 
However, unintentional blur on subjects can occur, often due to an incorrect focus point or an excessively large aperture.
Besides the real photos, image-generative models today also face problems in DoF.
These models allow users to create images from a text prompt~\cite{ho2020denoising, Rombach_2022_CVPR, esser2024scaling, songscore}. However, these models are trained on image-caption pairs, in which the captions rarely provide explicit descriptions of focus patterns or DoF characteristics. 
Also, it's hard to describe the focus points explicitly using only text.
As a result, prompt engineering alone is often insufficient for achieving precise control over focus placement and bokeh levels in generating images.

The above problems raise a natural question: \textit{can we arbitrarily refocus on an image that has already been captured or generated?} By selecting a focus point on the image and adjusting the blur level, akin to modifying the aperture and focusing by tapping on a camera screen, a refocus model could enable precise refocusing of an image, no matter what the original DoF effect is.

Motivated by this, we present \method, a diffusion-transformer-based refocus model that enables refocusing on an image by specifying an arbitrary focus point and a blur level, as shown in \cref{fig:teaser}.
However, training the model requires data annotated with specific focus points and blur levels of the same scenes, which are hard to collect.
To overcome the scarcity of such annotations, we design a simulation-based data collection pipeline to automatically render large-scale images focused on varied focus planes and blur levels from an all-in-focus image. These image pairs, derived from the same scene, enable the model to learn to generate a target image from a reference image under specified camera conditions.
With the data, we find that training on a single diffusion objective is insufficient to ensure precise refocus conditioning due to the complexity of refocusing.
Therefore, we propose a stacking constraint beyond the diffusion objective, which is an additional regularization term grounded in photographic focus stacking: photos focused on different focus planes can be linearly blended into a multi-focus image.
This constraint enforces DoF consistency across focus planes and blurriness during training, improving the precision and adherence to camera conditions in \method’s outputs.
We also design a depth dropout mechanism during training, which mitigates the model's over-reliance on depth maps, enabling the model to surpass the limitations imposed by the inaccuracies of the depth map. We also construct a benchmark of 150 scenes to evaluate the model.

Extensive experiments prove that \method supports robust refocusing on any subject within the image with varying bokeh levels regardless of its position or initial blurriness, providing highly flexible control over DoF adjustments for image post-processing.

\section{Related Work}

\noindent \textbf{Image Refocusing.} 
Post-capture refocusing has been extensively studied in computational photography, including methods based on specialized hardware—such as light-field cameras~\cite{ng2005light, ng2005fourier} or focal sweep cameras~\cite{zhou2012focal}—to acquire additional scene information, and can produce high-quality refocus results given the additional scene information.
However, these approaches rely on special hardware-captured data, and thus are not applicable when only a single RGB image is available as input.
Therefore, works~\cite{bando2007towards, zhang2011single} 
study on refocusing on a single image using deconvolution, but they produce obvious ring artifacts around edges.
RefocusGAN~\cite{sakurikar2018refocusgan}
uses GAN~\cite{goodfellow2020generative}
to refocus, but it can't modify the Bokeh level and supports low resolution.

\noindent \textbf{Simulating bokeh on an image.}
Bokeh rendering aims at simulating bokeh effects on all-in-focus images based on their depth map.
Classic rendering methods simulate the blur effect by estimating the blur level of each pixel based on its location on different depth layers using different algorithms~\cite{barron2015fast, hach2015cinematic, busam2019sterefo, luo2020bokeh, wadhwa2018synthetic, zhang2019synthetic}. However, these methods suffer from color leakage and artifacts from the boundaries of depth discontinuities. 
Neural-network-based rendering methods implicitly simulate the blur calculation to avoid artifacts by learning from image statistics~\cite{xiao2018deepfocus, lijun2018deeplens, dutta2021stacked}. But they are not as flexible as the classic ones due to the lack of controllability.
BokehMe~\cite{peng2022bokehme} integrates classic and neural rendering techniques into a hybrid framework to avoid artifacts and preserve flexibility.

However, these methods are limited to simulating bokeh effects on only all-in-focus images, which are not easy to acquire. Moreover, these methods heavily rely on the image's depth map, which can lead to inaccuracies if the predicted depth map is unreliable.

\noindent \textbf{Camera-conditioned diffusion models.}
Most works that condition diffusion models on cameras focus on the extrinsic parameters, \textit{i.e.,} the position and the rotation of the camera~\cite{hecameractrl, xing2025motioncanvas}. 
There are some works exploring controlling the intrinsic parameters of diffusion models.
Generative Photography~\cite{yuan2025generative} and Camera Settings as Tokens~\cite{fang2024camera} encode camera settings into a text-to-video/image model to control the DoF. However, they can only generate images from text, rather than editing existing images.
Bokeh Diffusion~\cite{fortes2025bokeh} allows editing the bokeh level of a reference image. However, the focused subject is implicitly decided by the model, limiting the flexibility in choosing focus points. In contrast, our method allows choosing arbitrary focus points and bokeh levels.
\section{Method}

\begin{figure*}[t]
    \centering
    \resizebox{\linewidth}{!}{
    \includegraphics{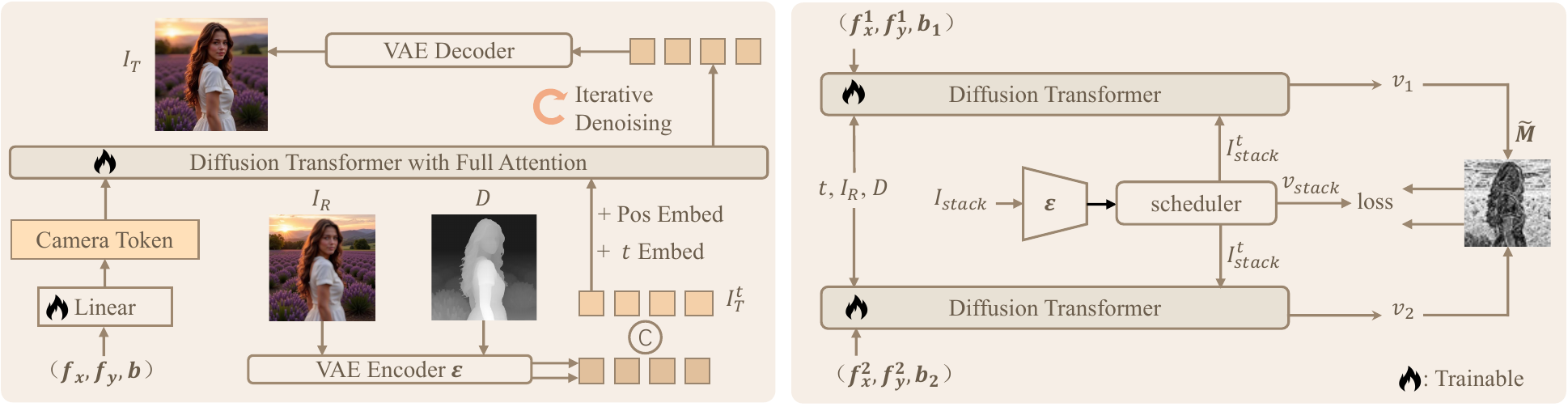}
    }
     \caption{\textbf{Pipeline of \method}. We convert the image and camera conditions into tokens using a VAE encoder or a learnable linear projection and input them into a diffusion transformer as shown on the left side.
     The right side visualizes the learning objective of the stacking constraint, where the two diffusion transformers share the same weights. The VAEs are all frozen and the diffusion transformer is trainable. The meaning of the symbols can be found at \cref{formulation: 1}.
     }
    
    \label{fig: pipeline}
\end{figure*}

To train a refocus model, we first collect the training data by simulating DoF (bokeh) pairs. Then, we use the simulated DoF images to train a diffusion transformer with our proposed stacking constraint.
\subsection{Simulating DoF Pairs}
\label{sec: simulate}
\noindent \textbf{Collect all-in-focus images}.
To train a refocus model conditioned on the reference image and different bokeh levels and focus points, we require bokeh pairs, \textit{i.e.,} image pairs capturing the same scene but with varying focus points and blur levels.
Acquiring such pairs directly from real-world photography is challenging due to the difficulty of maintaining consistent scene conditions across multiple shots with different camera parameters. Even slight changes in objects or lighting can introduce unwanted variations, making it impractical to capture perfectly aligned image pairs. To address this, we propose a simulation-based approach to simulate bokeh pairs from all-in-focus images. By rendering images with controlled DoF effects, we can systematically vary focus points and blur intensities while preserving scene consistency. 
We collect three types of all-in-focus images, including real-world photographs, phone-captured photos, and AI-generated images, to ensure quantity and diversity. We further discuss the details of dataset collection in the appendix.

\noindent\textbf{Creating DoF pairs.}
With the all-in-focus image, we can create the desired bokeh effects based on a target focus depth plane and blur level when an accurate depth map of the image is provided. 
Therefore, we employ Depth Anything V2~\cite{depth_anything_v2}, a monocular depth estimation model, to infer depth map $D$ for each image. 
For each all-in-focus image, we generate multiple bokeh pairs by systematically varying the focus depth $d$ and bokeh level $b$ using the predicted depth map and a simulation engine BokehMe~\cite{peng2022bokehme}. 
A single focus depth $d$ can give rise to multiple focus points $(f_x,f_y)$, defined as the set of pixels satisfying $||D(f_x, f_y)-d|| < \epsilon$, where $\epsilon$ is a small threshold representing the depth range considered in focus. This formulation represents that regions at approximately the same depth plane appear sharp simultaneously, mimicking the behavior of a real camera lens.

\subsection{Learning to Refocus through simulated DoF}
After constructing the simulated DoF images, we use a diffusion transformer~\cite{peebles2023scalable} to let the model learn to refocus through the image pairs. We also propose the stacking constraint and depth dropout to make the model more robust.

\noindent\textbf{Model Structure.}
After the simulation, we get groups of DoF images where each group has an all-in-focus image and various bokeh images simulated from this all-in-focus one, together with the focus plane and bokeh level annotations.
From each group in the dataset, we randomly pick one image as a reference image $I_R$ and another one as a target image $I_T$ as a data pair.
The learning target is to perform refocusing on $I_R$ given a new focus point $(f_x,f_y)$ with a new bokeh level $b$ to get the refocused image $I_T$. 
We follow the definition of $b$ in BokehMe.
Note that $I_T$ is conditioned on $(f_x,f_y)$ and $b$.
We also input the predicted depth map $D$ to the model, as the bokeh~(DoF) effects are related to the depth of the scene.
Our diffusion transformer then learns a network $\delta$ to generate an image by predicting the velocity $v$ in rectified flow~\cite{liu2023flow}:
\begin{align}
    \mathcal{L}_\text{flow} = || v -  \delta(I_T^t, t,f_x,f_y,b,I_R, D)||,
    \label{formulation: 1}
\end{align}
where $t$ is the flow matching time step and $I_T^t$ is the noise latent.

As illustrated in \cref{fig: pipeline}, we convert the input into tokens to feed to a diffusion transformer with full attention~\cite{chen2024unireal}. For image-level input, including condition images $I_R$, $D$, and the noise latent $I_R^t$, we encode them separately into latent space by a VAE encoder~\cite{kingma2013auto} and flatten them into a 1-D sequence of tokens.
For scalar input, including the camera parameters $(f_x, f_y, b)$, we project them into a camera token through a learnable linear projector $f:\mathbb{R}^3 \rightarrow \mathbb{R}^{d_{emb}}$, where ${d_{emb}}$ is the dimension of the attention embedding space.
The position embedding of the transformer is added to all tokens, and the timestep embedding of the diffusion algorithm is added to the noisy latent tokens $I_T^t$.
We concatenate all these tokens into a 1-D tensor sequence to feed to the transformer. The transformer utilizes the full attention mechanism~\cite{vaswani2017attention} to model the relationship between camera parameters and the image. During inference, the transformer iteratively denoises the noise latent into a denoised latent, which is decoded by the VAE decoder into the result-refocused image.

\noindent\textbf{Stacking constraint.}
A vanilla diffusion objective is insufficient to constrain the model to generate precise DoF effects. With only this single objective in training, the model often produces incorrect DoF behaviors such as blur or focus point mismatch in the result, since the DoF modeling is complex.
Inspired by the focus stacking in photography, we introduce the stacking constraint, an additional regularization term inspired by the focus stacking technique in photography to enhance DoF modeling.
Focus stacking linearly blends different images focused on distinct focal planes in the same scene into a multi-focus image using a mask $M$ marking the sharpest pixels of each image.
Specifically, given two images $I_1$ and $I_2$ focused on different focal planes in the same scene, the stacked image $I_\text{stack}$ is computed as:
\begin{align}
    M \odot I_1 + (1-M) \odot I_2 = I_\text{stack},
    \label{formulation: 2}
\end{align}
where $\odot$ denotes element-wise multiplication and the binary mask $M$ marks the sharper pixels of $I_1$ compared to $I_2$.
From \cref{formulation: 2}, we can derive that $M \odot I_\text{stack} = M \odot I_1$ and $ (1-M) \odot I_\text{stack} = (1-M) \odot I_2$, which ensures that the stacked image preserves the sharpest regions of each input image.
In the context of the rectified flow, a noise-perturbed image is defined as $I^t = I + vt$, where $v$ is the velocity and $t$ is the time step. We perturb $I_\text{stack}, I_1, I_2$ to get $I_\text{stack}^t,I_1^t,I_2^t$ with velocities $v_\text{stack},v_1,v_2$. 
The idea of the stacking constraint is to maintain the focus stacking relationship in \cref{formulation: 2} in the diffusion training to constrain the model to generate the DoF effects correctly and consistently across different variants.
Therefore, we enforce:
\begin{align}
    M \odot I_1^t + (1-M) \odot I_2^t = I_\text{stack}^t,
    \label{formulation: 3}
\end{align}
By substituting $I_t = I + v t$ into \cref{formulation: 3} and leveraging \cref{formulation: 2}, we reformulate the focus stacking formulation in terms of the diffusion prediction target $v$, yielding:
\begin{align}
    v_\text{stack} = M\odot v_1 +  (1-M)\odot v_2,
\end{align}
With this equation and the network $\delta$, the stacking constraint is:
\begin{align}
    \mathcal{L}_\text{stack} = || v_\text{stack} - (\widetilde{M}\odot \delta(I_\text{stack}^t, t, C_1) +  (1-\widetilde{M})\odot \delta(I_\text{stack}^t, t, C_2) )||,
    \label{formulation: 4}
\end{align}
where $\widetilde{M}$ is the down-sampled version of $M$ aligned with the latent space dimensions,
and $C_i = (f_x^i, f_y^i, b_i, I_R, D)$ is the camera conditions $(f_x^i, f_y^i,b_i)$ of image $I_i$ with the reference image $I_R$ and depth map $D$.
With the flow matching loss in \cref{formulation: 1}, the final loss is:
\begin{align}
    \mathcal{L} = \mathcal{L}_\text{flow} + \lambda \mathcal{L}_\text{stack},
    \label{formulation: stack}
\end{align}
where we use $\lambda =0.1$ as the default weight. We visualize this constraint on the right side of \cref{fig: pipeline}.

The stacking constraint enhances the refocus accuracy by imposing a physically grounded refocusing behavior that the two focusing results on the same image should be faithfully aligned with the scene structure and their corresponding focus points and bokeh levels respectively, 
because only in this way can they be stacked into the correct multi-focus image.
Note that the focus stacking in \cref{formulation: 2} and the stacking constraint in \cref{formulation: 4} can be extended to more than two images, but we use two images in practice.
Furthermore, the stacking constraint in \cref{formulation: 4} operates in the latent space with the mask $\widetilde{M}$ defined as a continuous-valued mask to enable soft blending of noise prediction targets.
More details about focus stacking and the stacking constraint are provided in the appendix.

\noindent\textbf{Depth dropout.}
Though we include the predicted relative depth maps during training to help understand the scene structure for better DoF modeling, it is suboptimal if the model heavily depends on these predicted depth maps due to their inaccuracy or ambiguity. Current SOTA bokeh-adding models like BokehMe will also be misled if the depth map is flawed.
For instance, when a transparent object (\textit{i.e.,} glass) appears in an image, current depth models predict the depth of the object’s surface rather than the occluded objects behind it. However, since light passes through the transparent object from the occluded objects, the DoF effect should be determined by the depth of the occluded objects, not the transparent object itself. Therefore, models that heavily depend on depth predictions would produce erroneous bokeh effects. Similarly, in complex scenes with unreliable depth maps, such dependency can mislead the model, compromising the quality of the refocused output.

To mitigate over-reliance on depth maps, we introduce a depth dropout mechanism that randomly drops 50\% of the depth maps during training by filling them with zeros. When depth maps are provided, the model learns the relationship among the image, depth, and bokeh effects. In the absence of depth input, the model is compelled to infer accurate bokeh effects independently of depth information. This approach enables \method to overcome the limitations imposed by the accuracy of depth maps due to over-dependence.

\subsection{Training Schemes}
During training, we randomly sample two images created in the simulation phase mentioned in \cref{sec: simulate} as a pair and use the training objective stated in \cref{formulation: stack} to supervise our model.
We also adaptively schedule the probability to balance the different training datasets: the photo dataset and the AI-synthetic dataset. At first, these two types of data are sampled with equal probability. As the optimization steps increase, we gradually increase the probability of AI-synthetic data and lower the probability of photos.
The reason is that web-crawled images contain more artifacts than high-quality AI-synthetic data, compromising the generation quality.
Thus, we adaptively increase the portion of AI-synthetic data to balance the training dataset for better quality.
\section{Experiments}
\subsection{Benchmark}
\label{sec: exp_benchmark}

\method enables arbitrary refocusing on a single image, a capability for which comparable work is scarce.
Therefore, we design multiple tasks and construct benchmarks correspondingly.

\noindent \textbf{(1) Refocus.}
The first task is arbitrary refocusing (the full task). 
We manually select two arbitrary focus points on an all-in-focus image and assign two bokeh levels to generate corresponding bokeh images, resulting in a pair of bokeh-rendered images.

\noindent \textbf{(2) Add bokeh.}
The second task is to add bokeh effects to only all-in-focus images. 
To construct the benchmark, we pick one focus point on an all-in-focus image and render four bokeh images of different bokeh levels using BokehMe.

\noindent \textbf{(3) Remove bokeh.}
The third task is to remove the bokeh effects from a bokeh image, \textit{i.e.,} deblur. 
We reuse the images created in the add-bokeh benchmark.

In total, we curated a benchmark dataset comprising 60 camera-captured photos, 30 phone-captured photos, and 60 AI-generated images. For each image, we construct two refocus samples, four bokeh samples, and four deblur samples.
This results in a benchmark of 150 scenes, each contains 10 samples.

\subsection{Metrics}
We calculate the MAE between the generated image and the simulated ground truth image to measure the accuracy of refocus. 
Generating clear content from blurred regions is inherently a multi-solution problem, as multiple plausible sharp representations may correspond to the same blurred input. Consequently, slight deviations between the generated results and the simulated ground truth are expected and acceptable.
To measure how the generated bokeh aligns with the target blur level, we follow Generative Photography~\cite{yuan2025generative} and Bokeh Diffusion~\cite{fortes2025bokeh} to calculate the average Laplacian variance trend for the generated bokeh images from an all-in-focus image, and then calculate the Pearson correlation with the reference bokeh level~(LVCorr). 
For evaluating scene consistency in refocusing, we employ the CLIP-I score~\cite{ruiz2023dreambooth} to measure semantic similarity between the generated and the reference image, leveraging semantic features encoded by the CLIP model~\cite{radford2021learning} to ensure that the generated image preserves the reference image’s content and context despite variations in DoF.
Furthermore, we directly examine the quality of the generated image without considering reference using CLIP Image Quality Assessment (CLIP-IQA)~\cite{wang2022exploring}.
For the bokeh removing sub-task, we adopt metrics in traditional deblur tasks, including MAE, LPIPS~\cite{zhang2018unreasonable}, and PSNR.

\subsection{Comparisons}

\begin{figure}[t]
    \centering
    \resizebox{\linewidth}{!}{
    \includegraphics{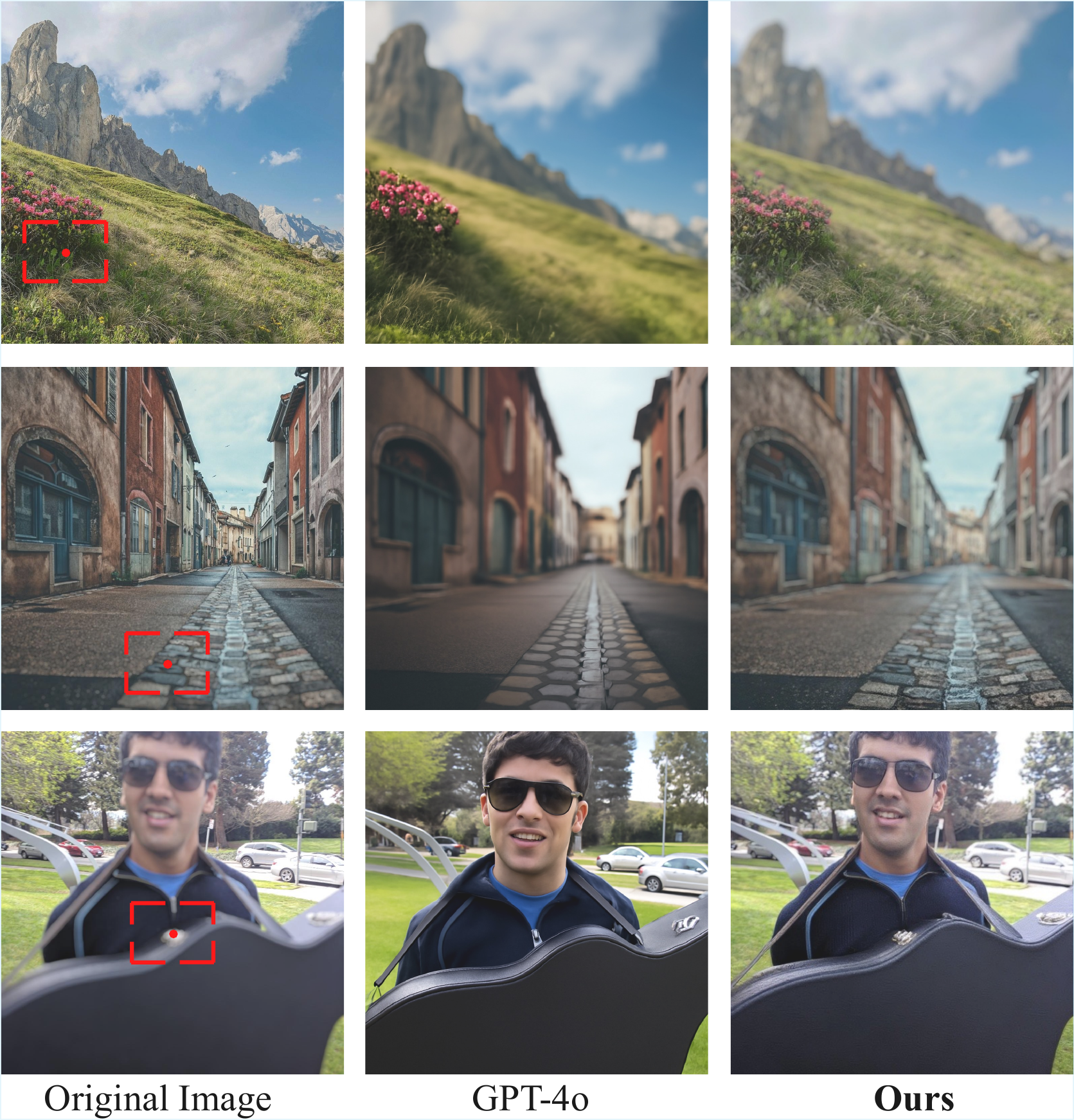}
    }
     \caption{\textbf{Qualitative comparisons on refocusing and adding bokeh.} We perform refocusing on images exhibiting strong defocus blur, setting the blur level to zero and fixing the focus point at the image center. 
     }
    \label{fig: comp_others}
\end{figure}

\begin{figure}[t]
    \centering
    \resizebox{\linewidth}{!}{
    \includegraphics{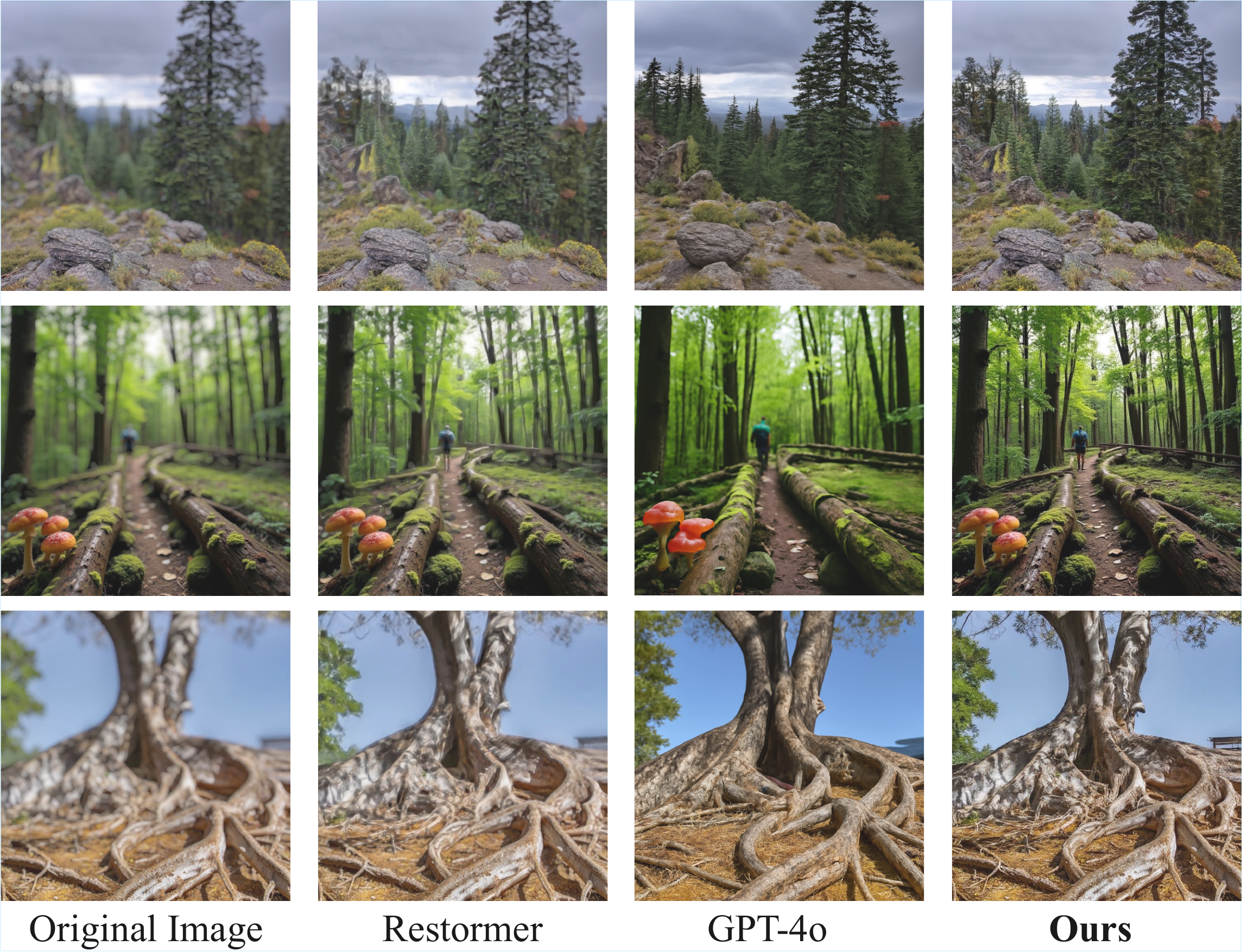}
    }
     \caption{\textbf{Qualitative comparisons on bokeh removing~(deblur).} We refocus on images with defocus blur, setting the blur level to zero, and fixing the focus point at the image center. We compare it with the SOTA deblur method Restormer and the image editing ability of GPT-4o.
     }
    \label{fig: comp_deblur}
\end{figure}

\begin{table}[!t]
    \centering
    \caption{\textbf{Quantitative comparisons on two tasks}. We compare \method with the image editing of GPT-4o on refocus and add bokeh.
    }
    \resizebox{!}{1.2cm}{
    \begin{tabular}{lcccccl}
    \toprule
    Sub-task & Method & CLIP-I~($\uparrow$) & MAE~($\downarrow$) & LVCorr~($\uparrow$) & CLIP-IQA \\
    \midrule
    \multirow{2}{*}{Refocus} & $\text{GPT-4o}^*$& 0.859 & 0.138& -- & 0.724 \\
     & \textbf{Ours} & \textbf{0.954} & \textbf{0.025} & -- & \textbf{0.834} \\
    \midrule
    \multirow{2}{*}{Bokeh} & $\text{GPT-4o}^*$ & 0.792 & 0.087 & -- & 0.567\\
     & \textbf{Ours} & \textbf{0.969} & \textbf{0.022 }& \textbf{0.920}& \textbf{0.857} \\
    \bottomrule
    \end{tabular}
    }
    \label{tab:quan_comp_refocus}
\end{table}

\begin{table}[!t]
    \centering
    \caption{\textbf{Quantitative comparisons on bokeh removing~(deblur).}  We report the deblur metrics in comparison with the SOTA deblur method Restormer and the image editing capacities of GPT-4o.}
    \resizebox{!}{0.9cm}{
    \begin{tabular}{lcccccl}
    \toprule
    Method & CLIP-I~($\uparrow$) & MAE~($\downarrow$) & LPIPS~($\downarrow$) & PSNR~($\uparrow$) & CLIP-IQA~($\uparrow$) \\
    \midrule
    $\text{GPT-4o}^*$ & 0.939 &0.149 & 0.583 & 13.736 & 0.611 \\
    Restormer & \textbf{0.971} & 0.044 & 0.674 & 24.120 & 0.857 \\
    \textbf{Ours} & 0.970 & \textbf{0.037} & \textbf{0.176} & \textbf{25.200} & \textbf{0.883}\\
    \bottomrule
    \end{tabular}
    }
    \label{tab:quan_comp_deblur}
\end{table}
We compare \method with other methods in different sub-tasks as described in \cref{sec: exp_benchmark} on the benchmark constructed by us. We can't compare with Bokeh Diffusion~\cite{fortes2025bokeh} because it's not open source, and it cannot specify the focus point explicitly.

\begin{figure*}[t]
    \centering
    \resizebox{\linewidth}{!}{
    \includegraphics{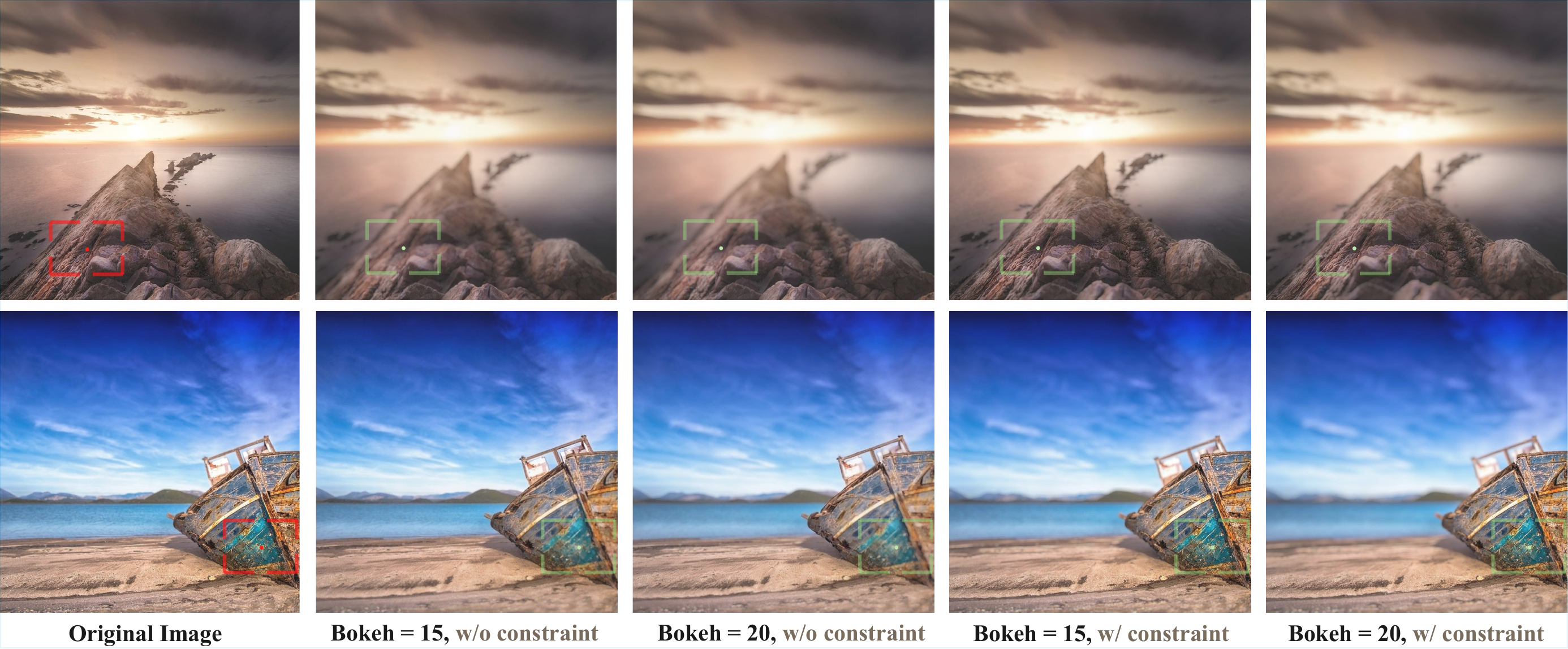}
    }
     \caption{\textbf{Qualitative ablation studies on the stacking constraint}. Without the stacking constraint, we observe incorrect model behaviors in generating bokeh effects: 
     in the first row, it fails to make the target stone area in focus; in the second row, the background is clear when bokeh=15, and most of the front part of the boat is blurry when bokeh=20, though it's in the focus plane.
     This shows that the stacking constraint enforces DoF condition consistency.
     }
    \label{fig: ablation_stack}
\end{figure*}

\begin{figure}[t]
    \centering
    \resizebox{\linewidth}{!}{
    \includegraphics{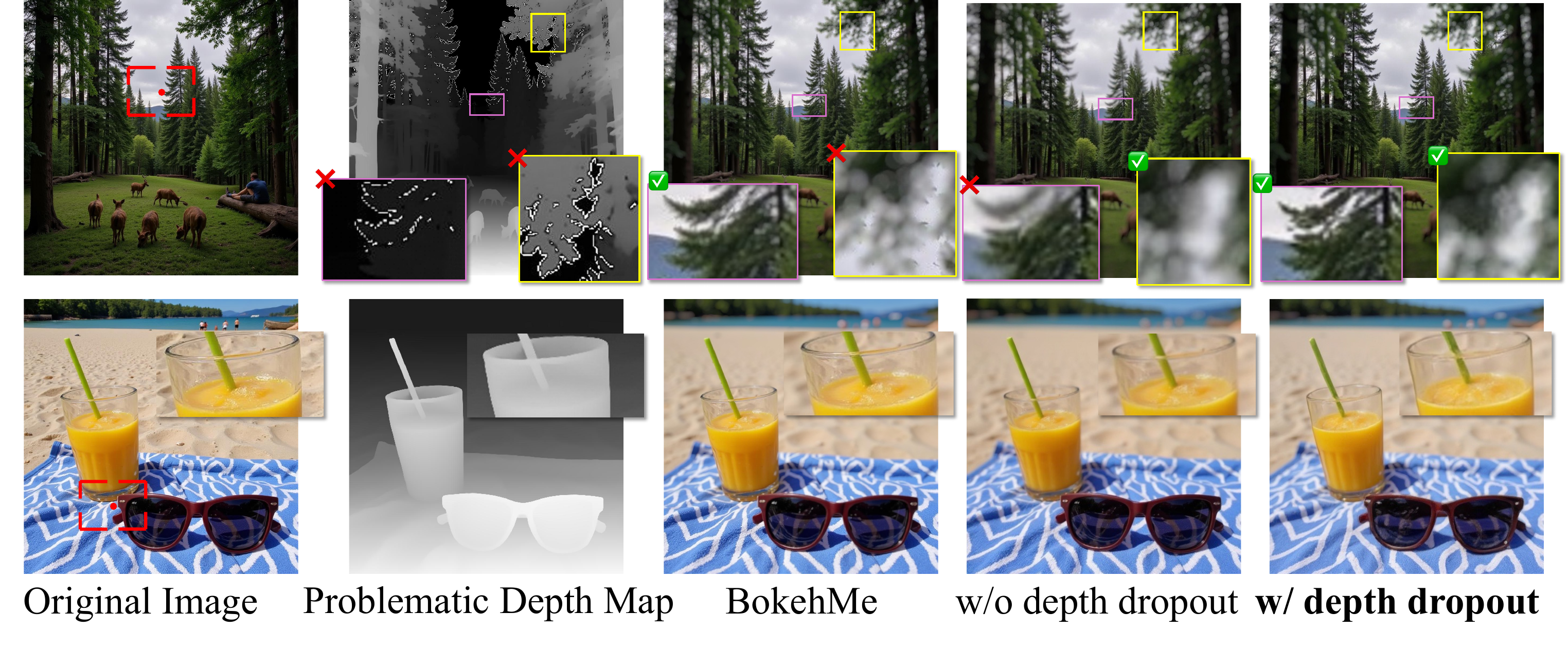}
    }
     \caption{\textbf{Qualitative studies on depth dropout.} We demonstrate that depth dropout enhances \method's robustness to inaccurate depth maps when simulating bokeh effects, outperforming the traditional bokeh simulator BokehMe and a variant of \method without depth dropout.
     }
    \label{fig: ablation_dropout}
\end{figure}

\noindent \textbf{Refocus.}
Due to the lack of similar works, we employ GPT-4o's image editing ability (\textit{i.e.,} DALLE3)~\cite{hurst2024gpt, betker2023improving}, which serves as a strong image editing model that supports editing an image by text prompting. We carefully design the prompt to let GPT-4o refocus on the targeted subject (but we can't control the target bokeh level quantitatively using only the text prompt). 
Due to resource limits, we cannot let GPT-4o perform all 1500 task samples of our benchmark. Therefore, we only choose 5\% of the task samples for GPT-4o. Combined with the qualitative study in \cref{fig: comp_others}, \method significantly surpasses GPT-4o on the refocus task, as GPT-4o struggles to preserve the scene consistency between the generated image and the reference image.

\noindent \textbf{Add bokeh.}
We prompt GPT-4o to add bokeh effects to all-in-focus images. 
Similar to the refocus task, \method achieves better consistency and accuracy compared to GPT-4o quantitatively in \cref{tab:quan_comp_refocus} and qualitatively in \cref{fig: comp_others}.
Note that we do not include the simulation-based method BokehMe in the qualitative study in \cref{tab:quan_comp_refocus} because it constructs the ground truth. 
Therefore, to demonstrate the drawbacks of traditional simulation-based methods in cases where the depth map is imperfect or ambiguous, we show these cases in the qualitative experiments \cref{fig: ablation_dropout}.
As shown in \cref{fig: ablation_dropout}, the depth dropout mechanism enhances \method's robustness to inaccurate or ambiguous depth maps, outperforming the traditional bokeh simulator BokehMe. In the first row, the depth map is inaccurate, making BokehMe simulate inconsistent DoF effects on these regions. In contrast, \method generates reasonable bokeh effects despite the inaccurate input depth map. In the second row, the depth map marks the depth of the glass instead of the sand behind, making BokehMe blur the region improperly. In contrast, \method successfully blurs the sand behind and keeps the glass sharp.

\noindent \textbf{Remove bokeh.}
\method is able to remove the bokeh effects on an image by assigning an arbitrary focus point and a bokeh level of 0. For GPT-4o, we prompt it to remove the bokeh effects. We also pick the SOTA deblur model Restormer~\cite{zamir2022restormer} for comparison.
As shown in the qualitative comparisons in \cref{fig: comp_deblur},
Restormer generates overly smooth content because it produces smooth, averaged results on ill-posed (overly blurred) regions as a regressive method.
GPT-4o cannot preserve the scene consistency from the reference image.
In contrast, \method generates sharp, reasonable, and consistent content for the blurred area, achieving superior deblur performances in the metrics in \cref{tab:quan_comp_deblur}. 

\subsection{Ablation Study}
\begin{table*}[!t]
  \centering
  \caption{\textbf{Quantitative ablation studies on data}. We conducted ablation studies on \method, focusing on data for training.
  We report average metrics across all sub-tasks, as well as specifically for the add-bokeh and remove-bokeh tasks.
  }
  \begin{tabular}{lccc|c|ccl}
    \toprule
    \multirow{2}{*}{Methods} & \multicolumn{3}{c}{All Sub-tasks} & \multicolumn{1}{c}{Add Bokeh} & \multicolumn{3}{c}{Remove Bokeh} \\
    \cmidrule(lr){2-4} \cmidrule(lr){5-5} \cmidrule(lr){6-8}
         & CLIP-I~($\uparrow$) & MAE~($\downarrow$) & CLIP-IQA~($\uparrow$)  &  LVCorr~($\uparrow$)  & MAE~($\downarrow$) &LPIPS~($\downarrow$) & PSNR~($\uparrow$) \\
    \midrule
    ground truth & 0.970 & 0.000  & 0.867 & 0.936 & 0.000 & 0.000 & +$\infty$ \\
    \midrule
    only photo & 0.953 & 0.034  & 0.842 & 0.781 & \underline{0.042} & 0.220 & \underline{24.283} \\
    only AI-synthesized & \underline{0.961} & \underline{0.032} & \underline{0.861} & \underline{0.856}&  0.043& 0.219 & 23.891  \\
    all data, w/o adaptive balancing & 0.961 & 0.035 & 0.852 & 0.824 & 0.042 & \underline{0.211} & 24.105 \\
    all data, w/ adaptive balancing & \textbf{0.966} & \textbf{0.029} & \textbf{0.863} & \textbf{0.920} & \textbf{0.037}  & \textbf{0.176} & \textbf{25.200} \\
    \bottomrule
  \end{tabular}
  \label{tab: quan_ablation_data}
\end{table*}

\begin{table*}[!t]
  \centering
  \caption{\textbf{Quantitative ablation studies on depth dropout}. We conducted ablation studies on \method, focusing on depth.
  }
  \begin{tabular}{lccc|c|ccl}
    \toprule
     \multirow{2}{*}{Methods} & \multicolumn{3}{c}{All Sub-tasks} & \multicolumn{1}{c}{Add Bokeh} & \multicolumn{3}{c}{Remove Bokeh} \\
    \cmidrule(lr){2-4} \cmidrule(lr){5-5} \cmidrule(lr){6-8}
         & CLIP-I~($\uparrow$) & MAE~($\downarrow$) & CLIP-IQA~($\uparrow$)  &  LVCorr~($\uparrow$)  & MAE~($\downarrow$) &LPIPS~($\downarrow$) & PSNR~($\uparrow$) \\
    \midrule
    w/o depth all the time & \underline{0.965} & 0.033 & 0.850  & 0.852 & 0.041 & 0.205 & 24.351 \\ 
    w/ depth, but w/o depth dropout & 0.960 & \underline{0.030} & 0.854 & 
    0.868 & 0.042 & 0.220 & 24.167 \\ 
    \textbf{w/ depth dropout} & \textbf{0.966} & \textbf{0.029} & \textbf{0.863} & \textbf{0.920} & \textbf{0.037}  & \textbf{0.176} & \underline{25.200} \\
    \textbf{w/ depth dropout} (0 depth for inference) & \textbf{0.966} & 0.032 & \underline{0.854} & \underline{0.878} & \underline{0.037} & \underline{0.176} & \textbf{25.219}\\
    \bottomrule
  \end{tabular}
  \label{tab: quan_ablation_depth}
\end{table*}

\begin{table*}[!t]
  \centering
  \caption{\textbf{Quantitative ablation studies on the stacking constraint}. We conducted ablation studies on \method, focusing on the stacking constraint.
  }
  \resizebox{0.905\textwidth}{!}{
  \begin{tabular}{lccc|c|ccl}
    \toprule
     \multirow{2}{*}{Methods} & \multicolumn{3}{c}{All Sub-tasks} & \multicolumn{1}{c}{Add Bokeh} & \multicolumn{3}{c}{Remove Bokeh} \\
    \cmidrule(lr){2-4} \cmidrule(lr){5-5} \cmidrule(lr){6-8}
         & CLIP-I~($\uparrow$) & MAE~($\downarrow$) & CLIP-IQA~($\uparrow$)  &  LVCorr~($\uparrow$)  & MAE~($\downarrow$) &LPIPS~($\downarrow$) & PSNR~($\uparrow$) \\
    \midrule
    w/o stacking constraint &0.962 & 0.030 & 0.861  & 0.858 & 0.041 & 0.200 & 24.350 \\
    \textbf{w/ stacking constraint} & \textbf{0.966} & \textbf{0.029} & \textbf{0.863} & \textbf{0.920} & \textbf{0.037}  & \textbf{0.176} & \textbf{25.200} \\
    \bottomrule
  \end{tabular}
  }
  \label{tab: quan_ablation_stack}
\end{table*}

We conduct a comprehensive ablation study of our method's core designs from the perspectives of data and training. We present average metrics across all sub-tasks and specifically for the bokeh-adding and bokeh-removing tasks.

\noindent \textbf{Training data.}
In \cref{tab: quan_ablation_data}, we compare \method trained under four data conditions: (1) solely on photographs, (2) solely on AI-synthesized images, (3) on all data with equal sampling probabilities, and (4) on all data with adaptive balancing.
Training exclusively on photographs yields the poorest performance across most tasks, except for deblurring, likely due to inherent noise and variability in web-sourced photographs.
Conversely, training solely on AI-synthesized images enhances refocusing precision and bokeh-adding consistency but degrades deblurring performance. A naive mixture of photographs and AI-generated synthetic images for training does not improve performance, yielding only moderate results across tasks.
With the adaptive balancing of the data distribution, \method achieves the highest performance across all tasks, likely due to the optimal integration of the complementary strengths of photos and AI-synthesized images.

\noindent \textbf{Depth dropout.}
In \cref{tab: quan_ablation_depth}, we evaluate the impact of depth maps on training \method under four conditions: (1) w/o depth maps, (2) w/ depth maps but w/o dropout~(\textit{i.e., }, training and inferring w/ depth input all the time), and (3) w/ depth maps and dropout, using depth input during inference, and (4) w/ depth maps and dropout, using empty depth input during inference. Omitting depth maps during training and inference significantly degrades performance across most tasks, except deblurring, as depth information is essential for generating accurate DoF effects. However, this condition outperforms training with depth maps without dropout in the deblurring task, where depth information is less critical for recovering clear content from blurred regions. Training with depth maps without dropout cannot get the optimal performance, for it strongly depends on the potentially inaccurate depth maps. In contrast, training with depth dropout achieves the highest performance, as it leverages depth information for accurate DoF effect generation while mitigating over-reliance, resulting in the most accurate and robust refocusing outcomes. We also observe that using empty depth maps (filled with zeros) as input for dropout-trained models only slightly degrades the performance, showing that \method exhibits robust performance despite the absence of depth information.

We further show that depth dropout enhances \method's robustness to inaccurate depth maps, outperforming training and inferring without depth dropout in \cref{fig: ablation_dropout}. Without depth dropout, the model strongly depends on the erroneous depth maps, leading to inconsistent DoF effects output for regions where the depth information is incorrect.
For example, it incorrectly blurs the background tree because of the flaws in the depth map.
In the second line, it incorrectly blurs the edge of the glass and does not blur the sand behind the glass correctly, similar to the behavior of BokehMe. In contrast, with depth dropout, the model produces correct results even when the input depth is inaccurate.
We further discuss the model's robustness against imperfect depth maps in the appendix. 

\noindent \textbf{Stacking constraint.}
We analyze the impact of the stacking constraint in \cref{tab: quan_ablation_stack}.
Stacking constraint significantly boosts model performance on all metrics across all sub-tasks, with particularly notable improvements in the LVCorr metric, which evaluates the accuracy of bokeh levels. This demonstrates that the stacking constraint effectively enforces focal plane and bokeh consistency, thereby improving the precision and adherence to specified focus and blur conditions in \method’s outputs.
Qualitative ablation studies are in \cref{fig: ablation_stack}. Without the stacking constraint, though the model can maintain the scene consistency, it generates inconsistent DoF effects (see the caption of \cref{fig: ablation_stack} for details).
Whereas the stacking constraint imposes correct DoF control on the model. 
\section{Conclusion}

In this paper, we propose \method, a method for training a diffusion-based refocus model that enables arbitrary refocusing of an image by specifying a focus point and bokeh level, delivering high-quality, robust, and consistent results.
This is achieved through large-scale training on simulated bokeh image pairs, imposing a photographically grounded stacking constraint, and a depth dropout mechanism. 
We construct a benchmark of 150 scenes to evaluate our refocus model.
Extensive quantitative and qualitative experiments demonstrate that the above approaches significantly improve refocusing precision and consistency across diverse scenes, even when depth information is inaccurate or unavailable.
All these results validate that our method offers unprecedented control over DoF adjustments, well-suited for real-world photo post-processing or generative AI applications. 
We also prepare vivid result demonstrations in the demo video in the supplementary materials.

\begin{acks}
This work is supported by the National Natural Science Foundation of China (No. 62422606, 62201484, 624B2124).
\end{acks}

\bibliographystyle{ACM-Reference-Format}
\bibliography{main}
\clearpage

\begin{figure*}[t]
    \centering
    \resizebox{0.96\linewidth}{!}{
    \includegraphics{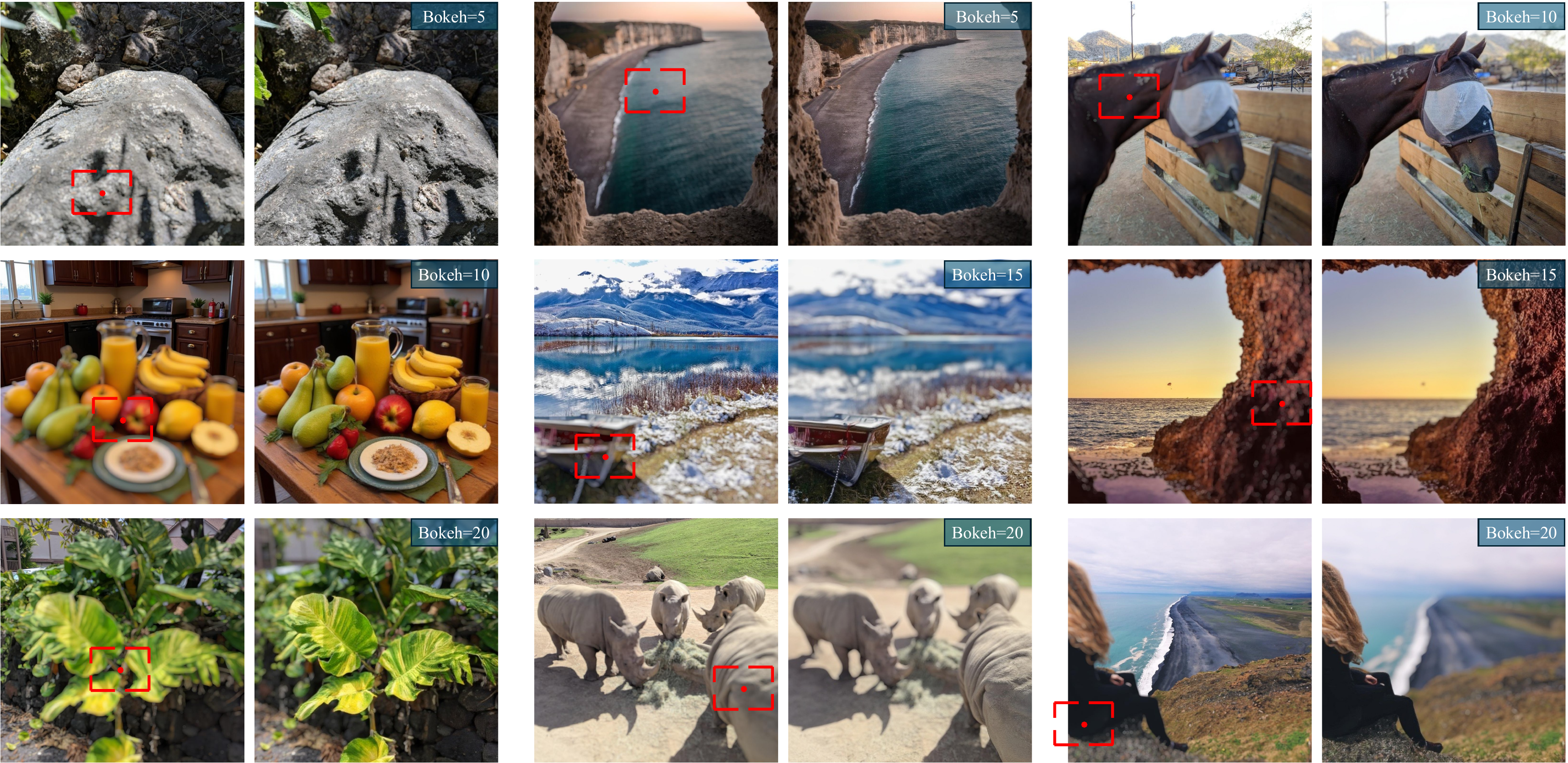}
    }
     \caption{\textbf{More visualizations on refocus on anything.} \method can refocus the existing image on an arbitrary focus point with a designated blur level, even if the focused subject is originally blurry.}
    \label{fig: refocus_1}
\end{figure*}

\begin{figure*}[t]
    \centering
    \resizebox{0.95\linewidth}{!}{
    \includegraphics{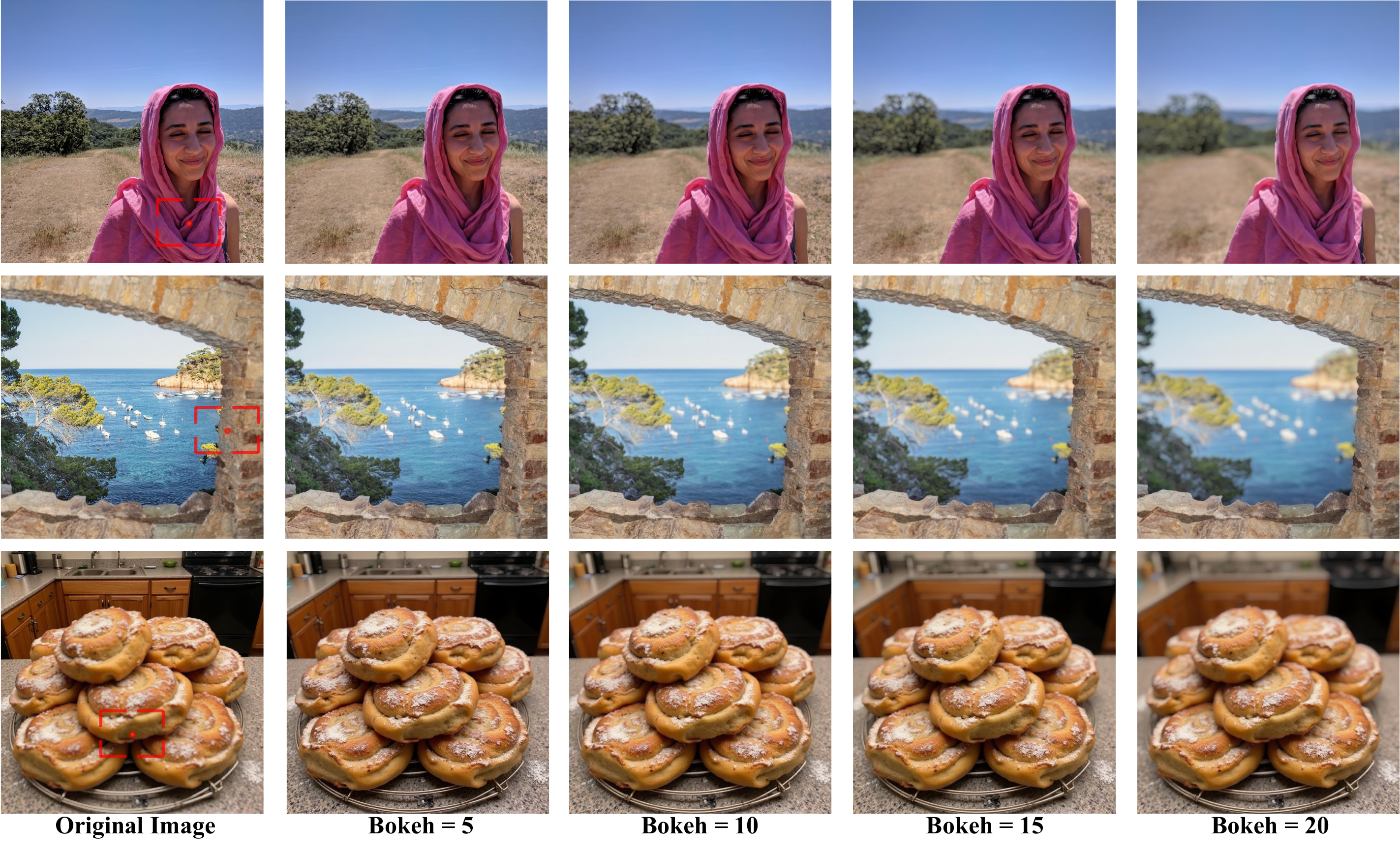}
    }
      \caption{\textbf{More visualizations on \method.} \method can refocus the existing image on an arbitrary focus point with different blur levels while maintaining high scene consistency across different blur levels.}
    \label{fig: refocus_2}
\end{figure*}

\begin{figure*}[t]
    \centering
    \resizebox{\linewidth}{!}{
    \includegraphics{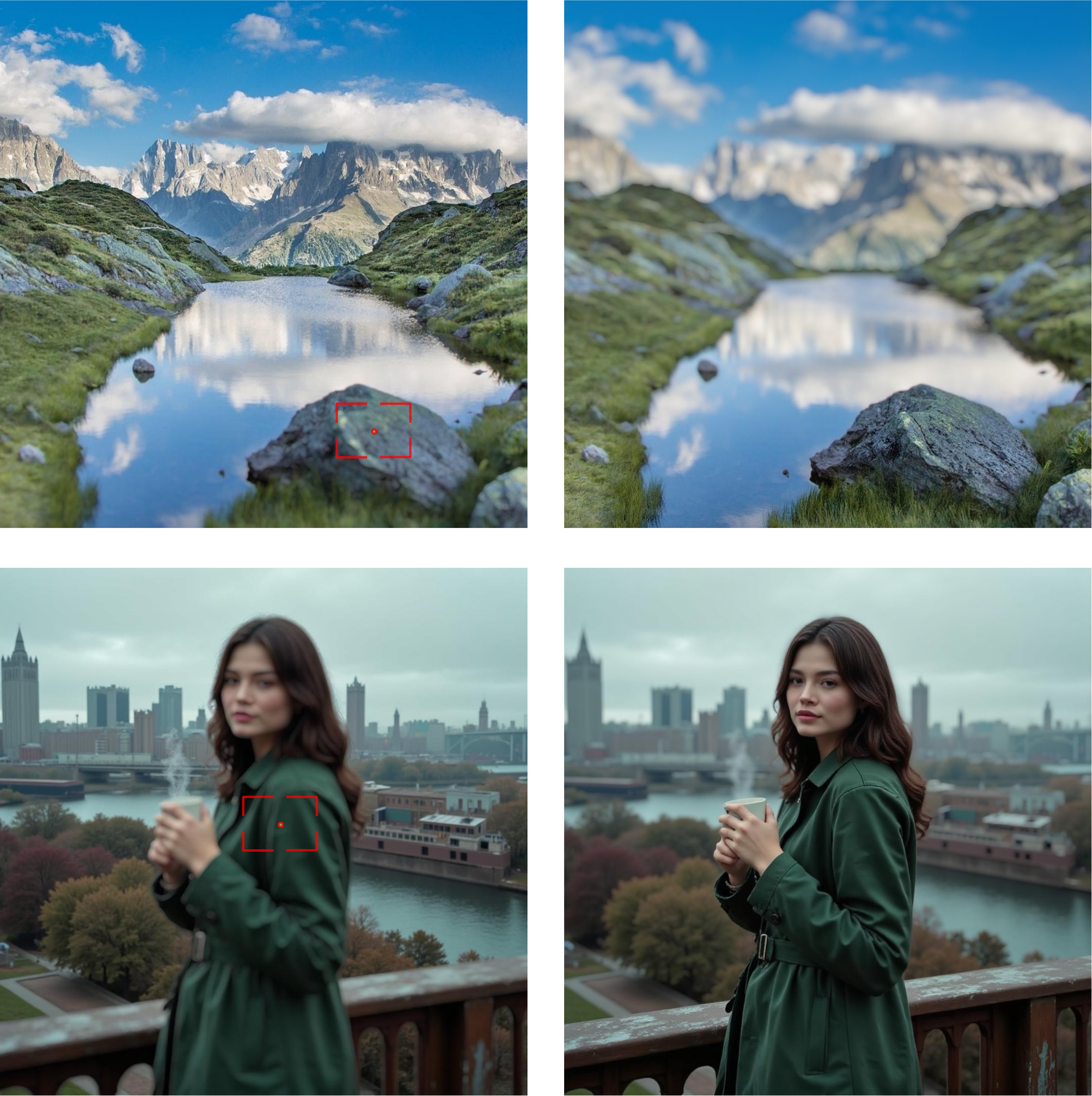}
    }
     \caption{\textbf{More visualizations on \method.} \method can refocus the existing image on an arbitrary focus point with different blur levels while maintaining high scene consistency across different blur levels on a resolution of $1024\times 1024$.}
    \label{fig: refocus_3}
\end{figure*}

\clearpage
\clearpage

\appendix
\section{Appendix Structure}

In the appendix, we first define bokeh level in \cref{appendix: bokehlevel}, and further analyze focus stacking and the stacking constraint in \cref{appendix: stack}.
Then, we discuss the implementation details of the dataset and training in \cref{appendix: train}.
After that, we further discuss the impacts of imperfect depth maps in \cref{appendix: imperfect depth}, and compare with more methods in \cref{appendix: more_comp}.
We also provide more qualitative results on real-world photos in \cref{appendix: real-world}.
Lastly, we discuss the limitations of \method in \cref{appendix: limitation}.

\begin{figure*}[htbp]
    \centering
    \resizebox{\linewidth}{!}{
    \includegraphics{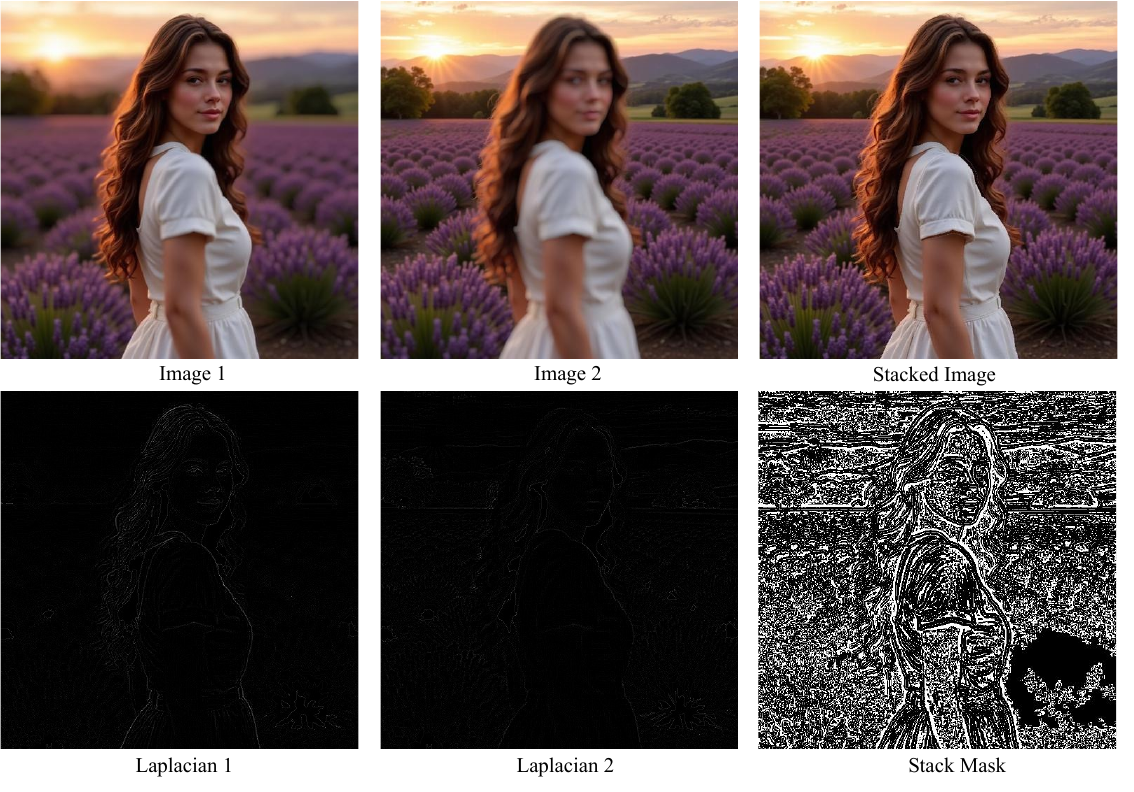}
    }
     \caption{\textbf{Illustration of focus stacking}. We stack image 1 and image 2 into the stacked image using the stack mask calculated from the Laplacian maps.
     }
    
    \label{fig: focusstack}
\end{figure*}

\section{Bokeh Level Definition}
\label{appendix: bokehlevel}

We follow BokehMe~\cite{peng2022bokehme} to define the bokeh level as:
\begin{align}
    b = \frac{r}{|d-d_f|},
\end{align}
where $r$ is the blur radius of a pixel, $d$ is the relative depth of this pixel, and $d_f$ is the relative depth of the focus plane (\textit{i.e.,} this image is focused on depth $d_f$).

\section{Focus Stacking and Stacking Constraint}
\label{appendix: stack}
\noindent \textbf{Focus Stacking.}
Focus stacking is a well-established technique, widely implemented in mobile phones and cameras, to estimate disparity, create all-in-focus photos, or synthesize depth-of-field for portrait photos.
~\cite{jacobs2012focal, kutulakos2009focal, suwajanakorn2015depth, wadhwa2018synthetic}.
However, this technique requires a focal stack, which is a set of images captured by sweeping the focal plane across the scene with the camera.
This requirement introduces two limitations: i) in many cases, users only have one image rather than a full focal stack, yet they may still wish to adjust the depth-of-field after capture; ii) in dynamic scenes, rapid motion during the focal sweep can lead to inconsistencies across images. This is why we introduce \method, a pure AI-based approach, taking only an RGB image as input, without requiring any other information, that can produce controllable refocusing results.

But we still gain insights from the focus stacking technique during model training, and we provide a sample of focus stacking in photography in \cref{fig: focusstack}. Image 1 and Image 2 show images focused on different focus planes while capturing the same scene. As a result, the human in image 1 is sharp, whereas image 2 is focused on the background. We then calculate the Laplacian maps of the images. A higher Laplacian value represents a pixel that has higher sharpness. Afterward, we calculate the stack mask by making comparisons between the two Laplacian maps. If the absolute Laplacian value of the pixel of image 1 is bigger than that in image 2, we then mark it as 1 (or the white pixel) in the stack map. Similarly, the 0 value pixels in the stack map represent pixels in image 2 that are sharper than the corresponding pixels in image 1. Note that all the bokeh variants are simulated from the same all-in-focus image, so the corresponding pixels between image 1 and image 2 are just pixels at the same position. This is another advantage of simulation: we don't have to additionally match the key points of the two images, which is commonly used in real-world focus stacking that is applied to photos that might have slight variations. With the stack mask, we calculate the resulting stacked image using the following formulation:
\begin{align}
    I_\text{stack} = M \odot I_1 + (1-M) \odot I_2,
\end{align}
As we can see from \cref{fig: focusstack}, the result stacked image merges the sharpest areas of the two input images.

\noindent \textbf{Stacking Constraint.}
The stacking constraint in the main body is similar to the focus stacking example described above, but it is performed in the latent space of the VAE. Therefore, we down-sample the stack mask $M$ by bilinear interpolation. The result down-sampled mask $\widetilde{M}$ is a continuous value mask instead of the original binary mask.
Given a VAE encoder $\mathcal{E}$, image 1 and image 2 are encoded into the latent space as $\mathcal{E}(I_1), \mathcal{E}(I_2)$. The focus stacking in the latent space is calculated as:
\begin{align}
I_\text{stack}=
     \mathcal{D}[\widetilde{M} \odot \mathcal{E}(I_1) + (1-\widetilde{M}) \odot \mathcal{E}(I_2)],
\end{align}
where $\mathcal{D}$ is the VAE decoder. We empirically verify that the resulting stacked image using the above formulation in the latent space is almost the same as the stacked image in the RGB space. Based on this fact, we derive the final stacking constraint in the latent space ($v_\text{stack} = M\odot v_1 +  (1-M)\odot v_2,$) in the main body.

\section{Additional Implementation Details of the Dataset and Training}
\label{appendix: train}
\begin{figure*}[t]
    \centering
    \resizebox{0.9\linewidth}{!}{
    \includegraphics{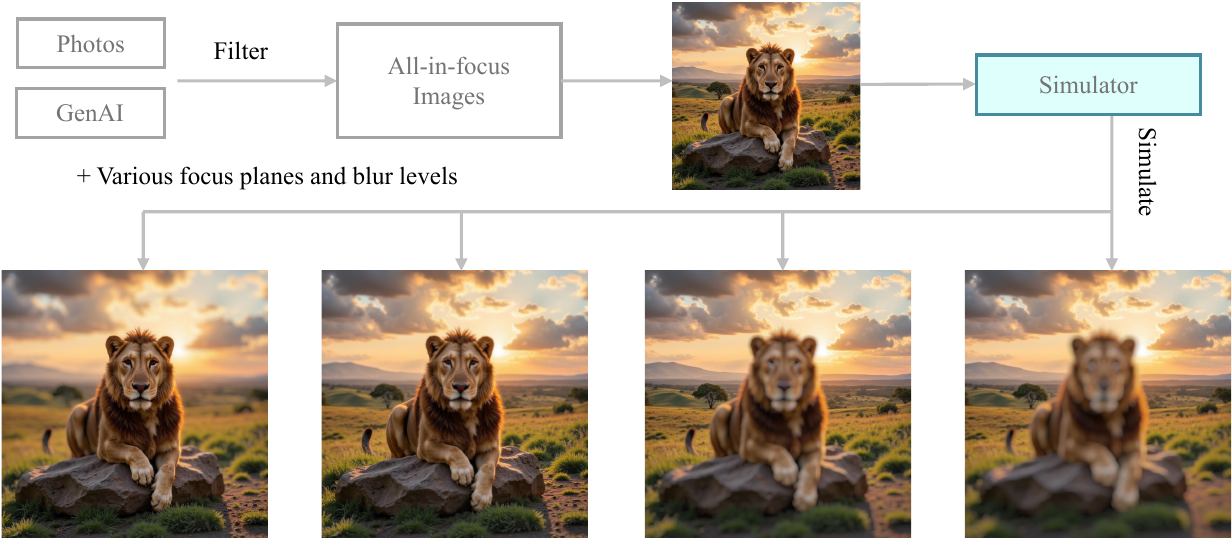}
    }
     \caption{\textbf{Data collection pipeline}. We simulate pairs on all-in-focus images with different focus planes and blur levels.}
    
    \label{fig: pipeline_data}
\end{figure*}

For the dataset, we simulate the bokeh pairs from all-in-focus images. We visualize the data collection pipeline in \cref{fig: pipeline_data}. As we can see, we start by collecting real-world photos and AI-synthesized images:

\begin{itemize}
\item \textbf{Real-world photographs}. Photos captured by cameras, particularly landscape images, are ideal due to their rich detail and depth. These images often leverage techniques like focus stacking, where multiple shots with different focus points are combined to produce an all-in-focus image.
Images during collection inevitably include bokeh. We filter these by analyzing sharpness across the image.

\item \textbf{Phone-captured photos}. Due to the small CMOS sensor size and compact lens design in smartphones, modern smartphones typically have a wide DoF, resulting in images where most objects appear sharp. We use the HDR+ Burst Photography Dataset~\cite{hasinoff2016burst}.

\item \textbf{AI-generated images}. Synthetic images created by advanced text-to-image models can be tailored to specific scenes, compositions, and lighting conditions, providing scalability and diversity in dataset design. We adopt a fine-tuned version (FLUX.1-dev-LoRA-AntiBlur LoRA) of FLUX.1-Dev~\cite{Flux}, which has been trained on a curated dataset of all-in-focus images to generate scenes with uniform sharpness across all depths.
\end{itemize}

We further filter the raw images and collect all-in-focus images by calculating and thresholding the Laplacian values of the images to filter out potentially blurred images. This is because higher Laplacian values represent higher sharpness, so we filter out images that have low Laplacian values. We further filter out images that may have bokeh effects by manual inspection.
After filtering, we get around 20k all-in-focus images of various scenes.

With the all-in-focus images, we use Depth Anything V2 to predict the depth map of each image. Afterward, we use BokehMe to simulate the bokeh effects on these all-in-focus images based on their corresponding depth maps.
The pixel value of the depth map ranges from $0$ to $1$, with $0$ standing for the farthest and $1$ representing the closest.
For each all-in-focus image, we iterate through 21 depth planes from $0.0$ to $1.0$ with a step length of $0.05$ as input focus planes for BokehMe. For each designated focus plane, we iterate through the bokeh levels from 1 to 20 to vary the level of blurriness. Therefore, a single all-in-focus image would produce $20\times 21 +1=421$ bokeh variants, resulting in $421\times421$ possible bokeh pairs.

For the backbone, our diffusion transformer utilizes FLUX.1-Dev, which has 11B parameters. To save training costs and preserve the original generation ability, we apply LoRA~\cite{hu2022lora} on the model. We use a LoRA rank of 64.
Starting from the base model FLUX.1-Dev, we fine-tune our models for 3600 optimization steps with a batch size of 256 and a resolution of $512\times512$. We adaptively adjust the sampling probabilities of data of different types. For the first 1000 steps, all the images are sampled with an equal chance. From 1000 to 2500 steps, we linearly adjust the sampling probability of AI-generated images to 100\% and lower photos to 0\%. 
After the first 3600 optimization steps in the resolution of $512\times 512$, we raise the training resolution to $1024\times 1024$ and continue to train for 3000 steps with a batch size of 128.
We train the model using a constant learning rate of $1e-4$ and 8 A100s.

\section{Impacts of Imperfect Depth Maps}
\label{appendix: imperfect depth}
In the main body, we introduce a depth-dropout technique to make the model more robust against depth input. 
It's also feasible to add different perturbations to the depth maps during training. But we choose only to use the dropout because it's the extreme case of perturbation (\textit{i.e., }data augmentation), and it's simple to implement.

Here we conduct a further analysis of \method's robustness against imperfect depth maps in \cref{tab: imperfect_depth}.
Apart from directly dropping out the depth map (\textit{i.e.,} filling the depth maps with zeros), we use different types of perturbations, including randomly masking the depth map (\textit{i.e.,} each pixel will be randomly set to 0 given a possibility, which we set as 30\%), randomly cropping out areas of the depth map(\textit{i.e.,} a random size rectangular regions in the depth map will be set to 0), and adding noise to the depth map (\textit{i.e.,} each pixel is added with a standard Gaussian noise independently).
The first two rows are the results that have already been in the main body in \cref{tab: quan_ablation_depth}.
As shown in the table, randomly masking or cropping regions of the depth maps only slightly decreases the performance of \method, indicating that our method is robust even when partial depth information is missing.
Perturbing the depth maps with Gaussian noise leads to a larger performance drop, suggesting that there is still room to improve robustness. This can be naturally addressed in future work by extending the depth dropout strategy to include fine-grained perturbations, \textit{i.e.}, randomly injecting noise into the depth maps during training.

\begin{figure*}[t]
    \centering
    \resizebox{\linewidth}{!}{
    \includegraphics{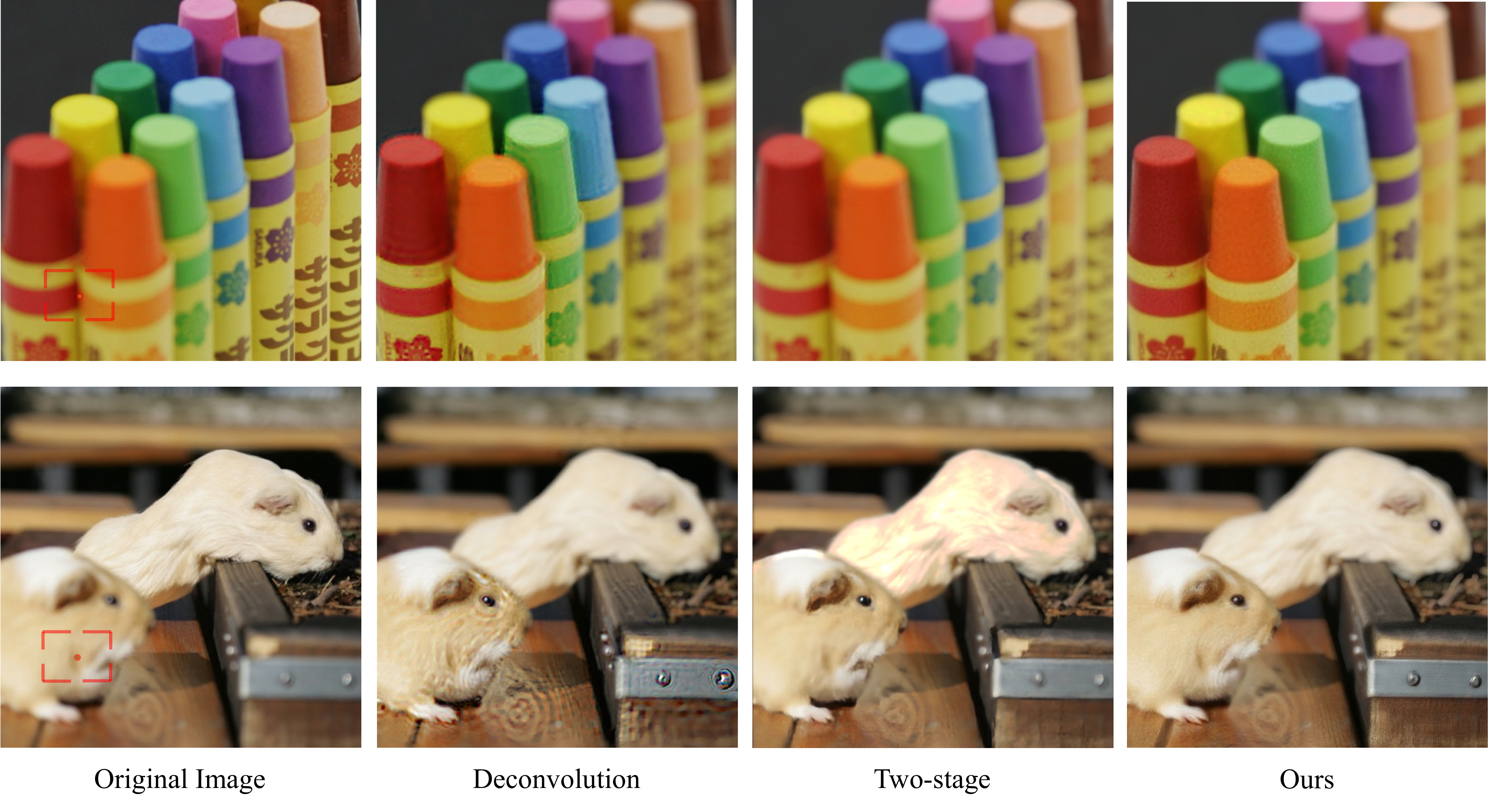}
    }
     \caption{\textbf{More comparisons.}. We compare \method with two representative refocusing methods. The first is a mathematics-driven approach that performs refocusing via deconvolution. The second is a hybrid pipeline that combines Restormer for deblurring with BokehMe for bokeh synthesis. }
    
    \label{fig: more_comp}
\end{figure*}

\begin{table*}[!t]
  \centering
  \caption{\textbf{Impacts of imperfect depth maps}. We conducted ablation studies on \method, focusing on imperfect depth.
  }
  \begin{tabular}{lccc|c|ccl}
    \toprule
   \multirow{2}{*}{Methods} & \multicolumn{3}{c}{All Sub-tasks} & \multicolumn{1}{c}{Add Bokeh} & \multicolumn{3}{c}{Remove Bokeh} \\
    \cmidrule(lr){2-4} \cmidrule(lr){5-5} \cmidrule(lr){6-8}
         & CLIP-I~($\uparrow$) & MAE~($\downarrow$) & CLIP-IQA~($\uparrow$)  &  LVCorr~($\uparrow$)  & MAE~($\downarrow$) &LPIPS~($\downarrow$) & PSNR~($\uparrow$) \\
    \midrule
    Full & \textbf{0.966} & \textbf{0.029} & \textbf{0.863} & \textbf{0.920} & \textbf{0.037}  & \textbf{0.176} & \underline{25.200} \\
    Full (0 depth for inference) & \textbf{0.966} & 0.032 & 0.854 & 0.878 & \underline{0.037} & \underline{0.176} & \textbf{25.219}\\
    Full (Random masking) & \underline{0.965} & \underline{0.029} & 0.852 & 0.916 & \underline{0.037} & 0.177 & 25.191 \\
     Full (Random cropping) & \textbf{0.966} &  \underline{0.029} &  \underline{0.856}& \underline{0.920}   & \underline{0.037} & \underline{0.176} & 25.197\\
     Full (Random noise) & \underline{0.965}  &  0.033 &  0.835&  0.901&  0.039 & 0.180 & 24.946\\
    \bottomrule
  \end{tabular}
  \label{tab: imperfect_depth}
\end{table*}

\section{More Comparisons}
\label{appendix: more_comp}
We further compare with two refocusing baselines.
The first is the work \textit{Towards Digital Refocusing from a Single Photograph}~\cite{bando2007towards}, which is a mathematics-driven method that performs refocusing via deconvolution.
The second method is a hybrid method that operates in two stages: it first applies a deblur model (\textit{i.e., } Restormer) to recover an all-in-focus image, and then uses a Bokeh simulator(\textit{i.e., } BokehMe) to simulate controllable bokeh effects on the all-in-focus image.

The qualitative comparison results are shown in \cref{fig: more_comp}. The deconvolution-based method produces noticeable ringing artifacts around high-contrast boundaries, such as the edge of the pen, the hamster’s fur, and the wooden texture. For the two-stage method, the overall quality is largely constrained by the deblurring stage. As shown in the figure, the first-row result remains blurred in the focus plane, while the second-row result exhibits overly smoothed patterns. In contrast, \method unifies the refocus problem into an end-to-end framework, succeeding in generating reasonable content in the blurry areas while accurately following the refocusing conditions.

\section{Results on Real-world Photos}
\label{appendix: real-world}
\begin{figure*}[t]
    \centering
    \resizebox{0.86\linewidth}{!}{
    \includegraphics{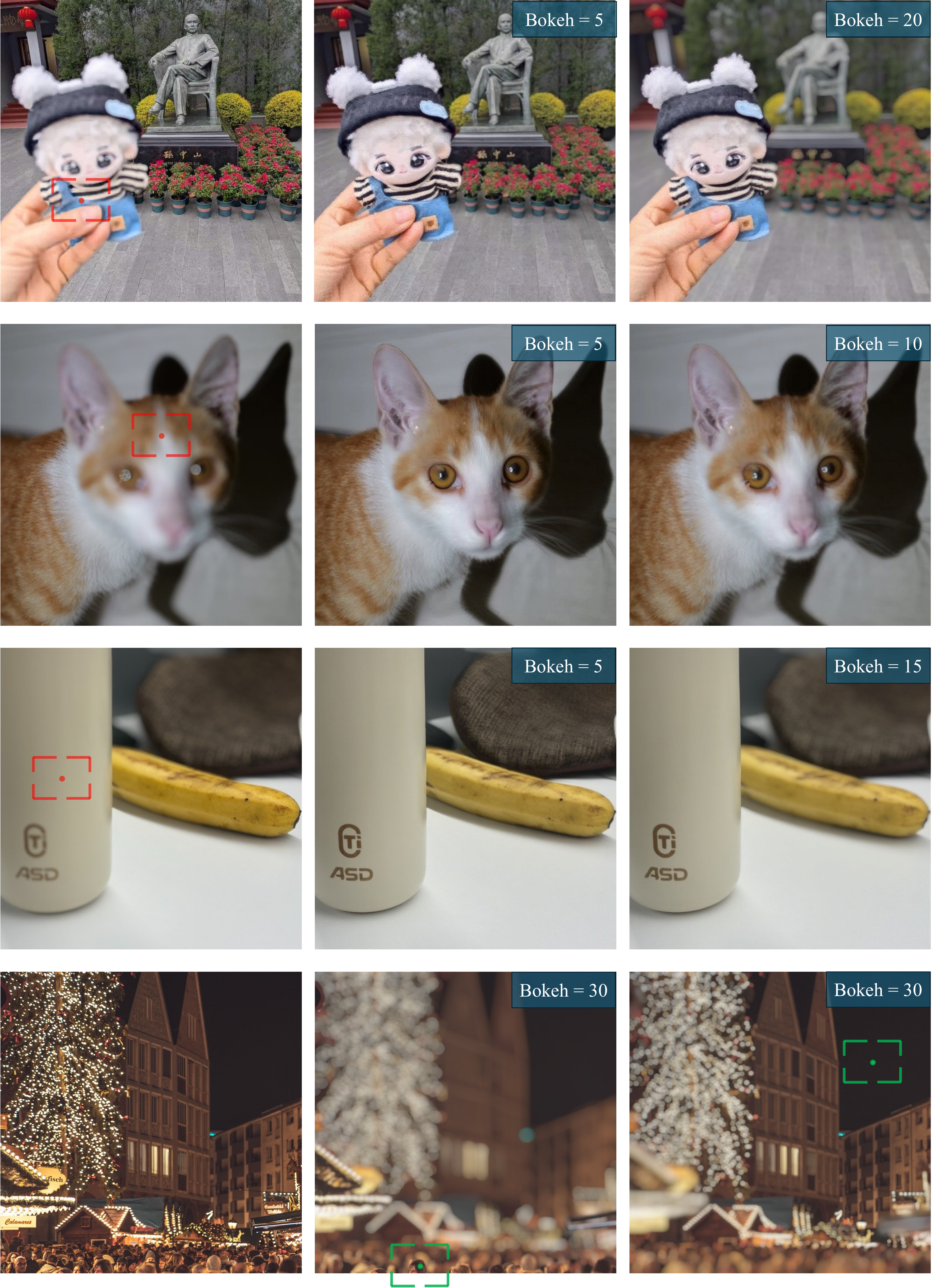}
    }
     \caption{\textbf{\method on real-world photos}. We apply \method on real-world photos with naturally occurring bokeh. }
    
    \label{fig: realworld}
\end{figure*}

We provide some refocus results on real-world photos using \method in \cref{fig: realworld}.
Photos from row 1 to row 3 are self-captured, and the photo in row 4 is from Pixabay.
These photos contain naturally occurring bokeh (blur), which allows us to assess \method's refocusing ability on real-world scenarios.
In the figure, the leftmost column shows the original photos that are input to the model. Photos from row 1 to row 3 are set to refocus on the red focus frames with different designated bokeh levels. Note that we manually set the red focus frames to the defocused areas. The qualitative results show that \method manages to refocus on the naturally blurred areas by generating reasonable content based on the original information, demonstrating its generalization ability on real-world photos.
The original photo in row 4 is nearly an all-in-focus photo. We designate the same bokeh level of 30 and set two different targeted focus points, which are marked by the green focus frames. The photo in the second column is refocused to the heads that are in a very close distance, while the photo in the third column is refocused to an infinite distance (\textit{i.e.,} the sky). The results show that \method can handle real-world scenarios with complicated object layouts. We admit that the bokeh balls are not as aesthetically pleasing as those in photos captured by professional camera sets, but styling the bokeh balls can be seamlessly integrated into our \method pipeline, which we will discuss in \cref{appendix: limitation}.

\section{Limitations and Discussions}
\label{appendix: limitation}

Though \method achieves high-quality refocusing on arbitrary focus points and different bokeh levels, it still has some limitations. 

Firstly, \method can only perform refocusing on images in the resolution of $512\times512$ or $1024\times1024$, which limits its ability to refocus on images of different aspect ratios (like 16:9) or higher resolution. This limitation can be solved in future studies by further scaling up the training in terms of resolution and the types of aspect ratios. 

Secondly, generating sharp content from blurriness is an ill-posed problem. Thus, the user may fail to recover their targeted content (\textit{e.g.,} one's portrait) for an overly blurred image. Though the synthesized content is reasonable, it is just one of the possible solutions for this multi-solution problem that doesn't match the user's expectations (\textit{e.g.,} the generated human is not similar to the user). This limitation can be solved by introducing another reference image (\textit{e.g.,} the user's other photo) to let the model perform refocus while generating the target sharp content by referring to a reference target.
Additionally, this ill-posed problem may lead to artifacts in areas requiring deblurring. This is due to the model capacity limitation of the original pretrained model backbone -- it may generate content with artifacts in complex cases. By leveraging a more powerful backbone and training on larger numbers of high-quality data, these artifacts can be mitigated in future studies.

Thirdly, \method can only specify the bokeh level and the targeted focus point, but cannot perform finer bokeh control, such as styling the bokeh shape.
However, finer control capacity can be seamlessly integrated into our pipeline in future studies. We take adjusting the bokeh shape as an example, which can be incorporated into \method through: 
1) Creating data of various bokeh shapes. This can be implemented using Bokeh simulators such as BokehMe, which treats the bokeh ball as a polygon, and the number of edges can be manually set.
2) Encoding a condition representing the bokeh shape into the camera token. For example, we can use a scalar to represent a specific type of bokeh shape and project it into the camera token, similar to the focus point coordinates and bokeh level.
3) Training the model with the above data and model.

\end{document}